\documentclass[10pt,twocolumn,letterpaper]{article}

\usepackage{iccv}
\usepackage{times}
\usepackage{epsfig}
\usepackage{graphicx}
\usepackage{bbm}
\usepackage{amsmath}
\usepackage{amssymb}
\usepackage{nicefrac}       
\usepackage{microtype}      
\usepackage{bm}
\usepackage{booktabs}
\usepackage{multirow}
\usepackage{color}
\usepackage{float}
\usepackage{enumitem}
\usepackage{subfig}
\usepackage{tabulary,multirow,overpic,xcolor}
\usepackage{caption}
\usepackage[utf8]{inputenc}

\newcolumntype{x}[1]{>{\centering\arraybackslash}p{#1pt}}

\newcommand{\app}{\raise.17ex\hbox{$\scriptstyle\sim$}}

\newcolumntype{x}[1]{>{\centering\arraybackslash}p{#1pt}}

\newlength\savewidth\newcommand\shline{\noalign{\global\savewidth\arrayrulewidth
  \global\arrayrulewidth 1pt}\hline\noalign{\global\arrayrulewidth\savewidth}}
\newcommand{\tablestyle}[2]{\setlength{\tabcolsep}{#1}\renewcommand{\arraystretch}{#2}\centering\footnotesize}

\definecolor{citecolor}{RGB}{34,139,34}
\usepackage[pagebackref=true,breaklinks=true,letterpaper=true,colorlinks, citecolor=citecolor,bookmarks=false]{hyperref}

\usepackage[pagebackref=true,breaklinks=true,letterpaper=true,colorlinks,bookmarks=false]{hyperref}

\iccvfinalcopy 

\def\iccvPaperID{****} 

\begin{document}

\title{Reasoning About Human-Object Interactions Through Dual Attention Networks}

\author{Tete Xiao$^{1,2}$\thanks{The work was done when Tete Xiao was an intern at IBM Research, Cambridge.} \quad Quanfu Fan$^2$ \quad Dan Gutfreund$^2$ \quad Mathew Monfort$^3$ \quad Aude Oliva$^3$ \quad Bolei Zhou$^4$
 \\ 
\and
$^1$University of California, Berkeley\\
\and
$^2$MIT-IBM Watson AI Lab, IBM Research\\
\and
$^3$Massachusetts Institute of Technology\\
\and
$^4$The Chinese University of Hong Kong\\
}
\maketitle

\begin{abstract}
Objects are entities we act upon, where the functionality of an object is determined by how we interact with it. In this work we propose a Dual Attention Network model which reasons about human-object interactions. The dual-attentional framework weights the important features for objects and actions respectively. As a result, the recognition of objects and actions mutually benefit each other. The proposed model shows competitive classification performance on the human-object interaction dataset Something-Something. Besides, it can perform weak spatiotemporal localization and affordance segmentation, despite being trained only with video-level labels. The model not only finds when an action is happening and which object is being manipulated, but also identifies which part of the object is being interacted with. Project page: \url{https://dual-attention-network.github.io/}.

\end{abstract}

\section{Introduction}

\begin{figure}[!t]
\centering
\includegraphics[width=.85\linewidth]{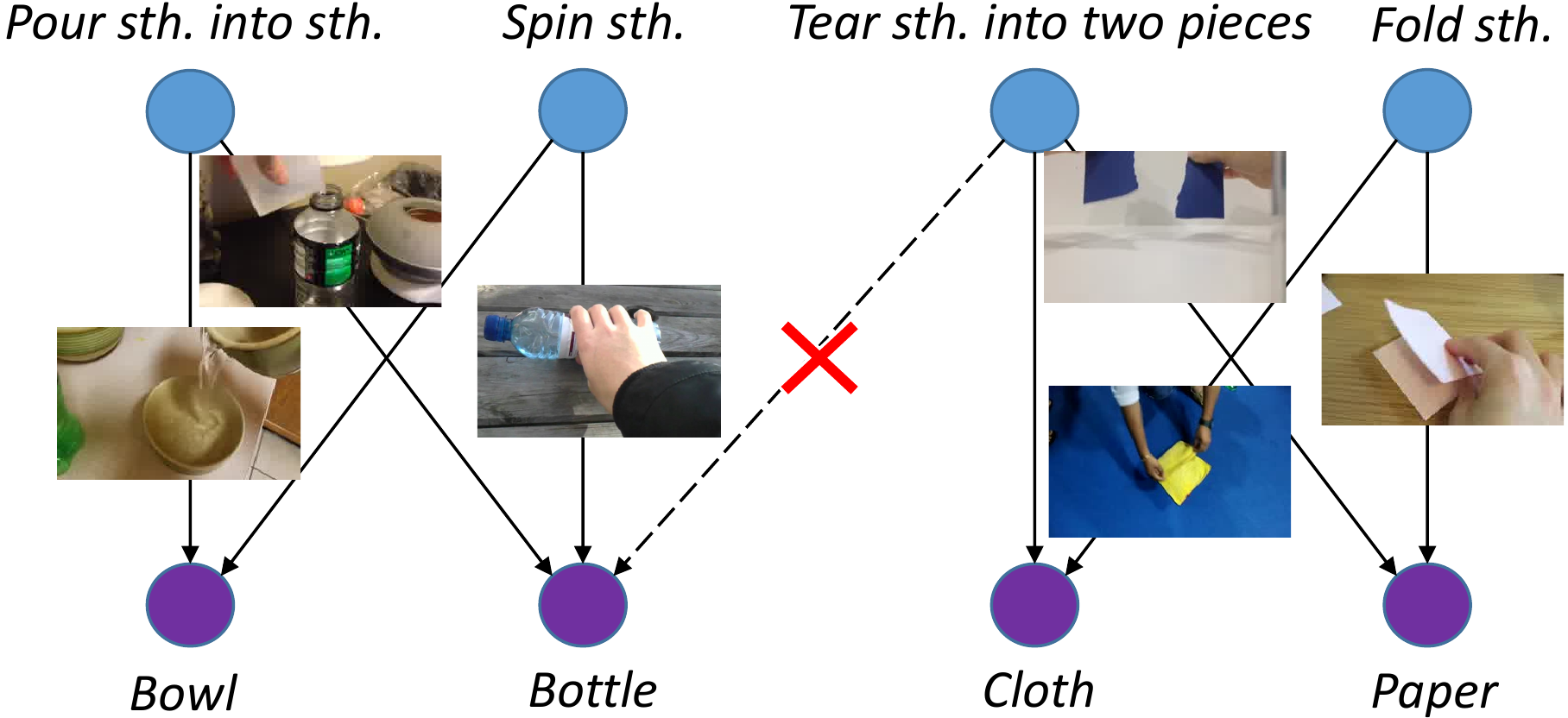}
\caption{\textbf{Object and action co-dependence}. The action \textit{tearing something into two pieces} can be performed on a piece of \textit{paper} but not a \textit{bottle}. Given the object \textit{bottle}, we can pour water into it or spin it, but cannot fold it or tear it.}
\vspace{-0.3cm}
\label{fig:teaser}
\end{figure}

Affordance, introduced by James Gibson~\cite{gibson1966senses}, refers to the properties of an object, often its shape and material, that dictate how the object should be manipulated or interacted with. The possible set of actions that an object can afford is constrained. For instance, we can drink from a plastic bottle, pour water into it, squeeze it, or spin it, but we cannot tear it easily into two pieces (see  Figure~\ref{fig:teaser}). Similarly, for a given action, the possible objects which it can apply to are also limited. For example, we can fold a paper but not a bottle.

A handful of works have exploited object information for the recognition of Human-Object Interactions (HOIs) and more general action recognition~\cite{delaitre2011learning,gkioxari2017detecting,qi2018learning, kalogeiton2017joint,Wang_2018_ECCV}. However, understanding HOIs goes beyond the perception of objects and actions: it involves reasoning about the relationships between how the action is portrayed and the consequence on the object (\ie, whether the shape or location of the object is changed by the action upon it). Most of the previous works pre-define human-object or action-object pairs for HOI~\cite{delaitre2011learning,gkioxari2017detecting,qi2018learning}. The classification is done by either a graphical model~\cite{gupta2009observing}, a classifier based on the appearance features~\cite{gkioxari2017detecting}, or a graph parsing model~\cite{qi2018learning}. A potential issue with the previous approaches is that the complexity of an HOI model grows quickly as the number of objects and actions increase. The reasoning capability of these approaches is also limited due to the action-object pairs being preset for modeling. In addition, full annotations including action labels and object bounding-boxes are often required by these approaches for the effective modeling of HOIs.

\begin{figure*}[!th]
\centering
\includegraphics[width=0.85\textwidth]{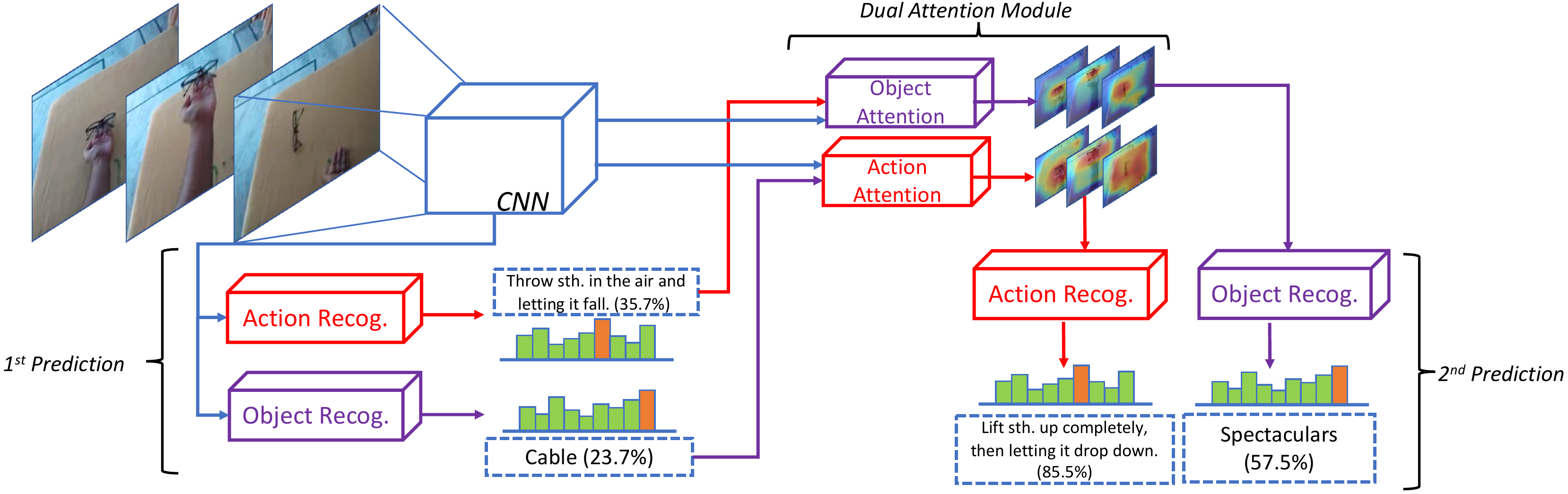}
\caption{\textbf{Framework overview}. Our approach exploits the role of human action and object in human-object interactions via the dual attention module. The Dual Attention Network first predicts plausible action and object labels independently as the priors ($1^{st}$ prediction). Then the priors are used to generate attention maps that weight the features of object and action for the $2^{nd}$ prediction. Action Recog.: action recognition head. Object Recog.: object recognition head.
}
\vspace{-0.2cm}
\label{fig:dual_attention_network}
\end{figure*}

Here we propose a Dual Attention Network model that leverages object priors as the guidance to where actions are likely to be performed in a video stream and \emph{vice versa}. The focus of attention is represented by a heatmap indicating the likelihood of where an action is taking place or where an object is being manipulated in each frame. These attention maps can enhance video representation and improve both action and object recognition, yielding very competitive performance on Something-something~\cite{goyal2017something} dataset. We show that the attention maps are intuitive and interpretable, enabling better video understanding and model diagnosis.  Such attention maps also facilitate weakly-supervised spatiotemporal localization of objects and actions.

\subsection{Related work}
{\bf Action recognition.} Deep convolutional neural networks have been used with success for action recognition ~\cite{krizhevsky2012imagenet,szegedy2015going,xie2017aggregated,he2016deep,Hu_2018_CVPR}. 
For instance, we can exploit the success of CNNs for static images and RNNs for temporal relations by feeding CNN-based features from single frame into an RNN model \cite{yue2015beyond,donahue2015long,karpathy2014large}. An alternative approach is to extend 2D CNNs by applying 3D convolutional filters (C3D) on raw videos to directly capture the spatiotemporal information~\cite{tran2015learning}. The 3D filters can be ``inflated'' from 2D filters (I3D)~\cite{carreira2017quo} and can be initialized with an ImageNet~\cite{deng2009imagenet} pre-trained model. Recent works involve Non-local Networks~\cite{Wang_2018_CVPR}, which uses space-time non-local operations to capture long-range dependencies; and Temporal Relation Network~\cite{Zhou_2018_ECCV}, which sparsely samples frames from different time segments and learns their causal relations. In addition to the end-to-end frameworks on raw video inputs, optical flow~\cite{horn1981determining} has also proven to be useful~\cite{simonyan2014two,carreira2017quo,Zhou_2018_ECCV} when combined with features extracted from raw RGB images.

{\bf Human-object interactions and visual affordance.} Several works have exploited human-object interactions and affordance for action recognition. Gupta \etal~\cite{gupta2009observing} integrate perceptual tasks to exploit the spatial and functional constrains for understanding human-object interactions. Koppula \etal~\cite{koppula2016anticipating} frame the problem as a graph, where the nodes represent objects and sub-activities while the edges represent the affordance and relations between human actions and objects. The graph model can be optimized using structural Support Vector Machine (SVM)~\cite{koppula2016anticipating} or Conditional Random Field (CRF)~\cite{lafferty2001conditional}. Jain \etal~\cite{jain2016structural} merge spatiotemporal graph with an RNN to model different kinds of spatial-temporal problems such as motion, action prediction and anticipation. Gkioxari \etal~\cite{gkioxari2017detecting} propose InteractNet to detect $\langle$ human, verb, object $\rangle$ triplets by exploiting the appearance features from detected persons. Dutta and Zielinska~\cite{dutta2017action} employ a probabilistic method to predict the next action in human-object interactions~\cite{wang2011action}. Fang \etal~\cite{Fang_2018_CVPR} propose a model to learn the interactive region and action label of an object via watching demonstration videos.

\begin{figure}[!t]
\begin{centering}
\includegraphics[width=.8\linewidth]{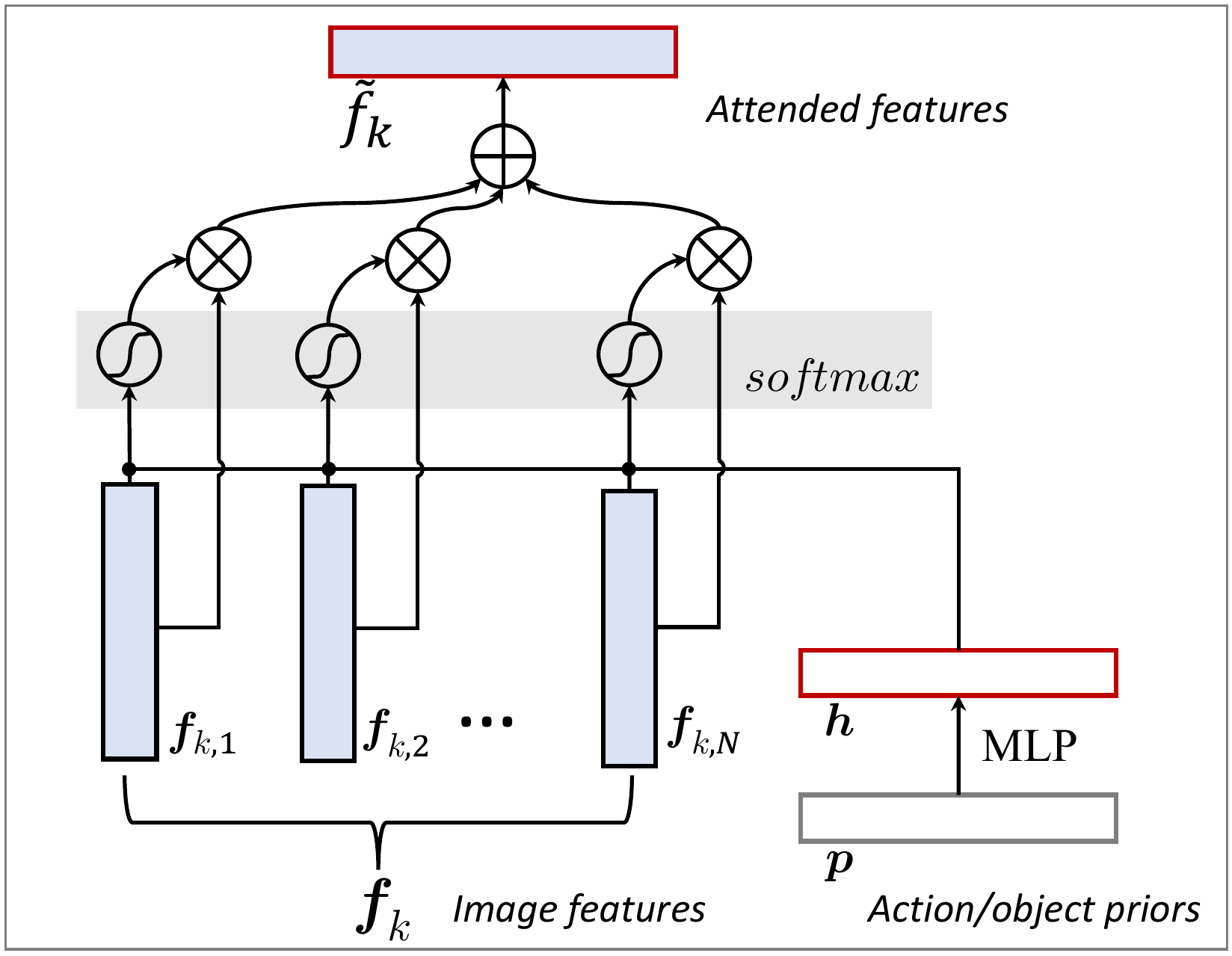}
\caption{\textbf{Illustration of the attention module} for the $k^{th}$ frame. It encodes action (object) priors and attends image regions accordingly, yielding the representation for object (action) recognition.}
\label{fig:att_block}
\end{centering}
\vspace{-0.1cm}
\end{figure}

{\bf Attention models.} Attention mechanism has been adopted for action recognition. Sharma \etal~\cite{sharma2015action} use a soft attention module to re-weight CNN features spatially. Ramanathan \etal~\cite{ramanathan2016detecting} propose to attend people involved in specific events for event detection in multi-person videos. Song \etal~\cite{song2017end} exploit skeleton data for attention module to extract more discriminative features in human-centered actions. Du \etal~\cite{Du2018RecurrentSA} propose to incorporate a spatial-temporal attention module into a classical CNN-RNN video recognition model. 

Co-attention models ~\cite{shih2016look,xiong2016dynamic,xu2016ask,yang2016stacked,lu2016hierarchical,lu2017knowing} are widely adopted in tasks relating to language and vision such as image captioning~\cite{vinyals2015show}, visual question answering (VQA)~\cite{antol2015vqa} and visual question generation (VQG)~\cite{mostafazadeh2016generating}. 
Lu \etal~\cite{lu2016hierarchical} propose a hierarchical co-attention model for VQA, in which image representation is used to guide the question attention and \emph{vice versa}, exploiting the relation between the two modalities, image and text. 

{\bf Comparison to our approach.}
In contrast to the self-attention and human-attention models for action recognition, and the co-attention models for multi-modal (text and vision) tasks, our framework applies dual attention in the context of multi-task learning on a single input modality, namely the raw video. Our novel iterative model exploits the action/object relations to simultaneously learn cross-task object/action attention maps, which significantly differs from previous works that use self-guided attention~\cite{sharma2015action,Du2018RecurrentSA}. Our model is able to not only outperform the previous state-of-the-art on a human-object interaction dataset~\cite{goyal2017something} but also yield interpretable attention maps (see Section \ref{sec:localization}).

\section{Dual Attention Network for Human-Object Interactions}

The dual attention network is designed in such a way that the streams of human activity and objects interact with each other by cross-weighting the intermediate features of action and object for recognition. Our attention module is general and can be plugged into any CNN-based action recognition models for feature enhancement. We first describe CNN-based feature representations for video understanding in Sections~\ref{sec:extract_video_representation} and~\ref{sec:object_recognition}. We then introduce the dual attention model in Section~\ref{sec:dual_attention_module}, the building block for reasoning about actions and objects. Finally we detail the full framework in Section~\ref{sec:full_framework}.

\begin{figure*}[!th]
\centering
\includegraphics[width=0.85\textwidth]{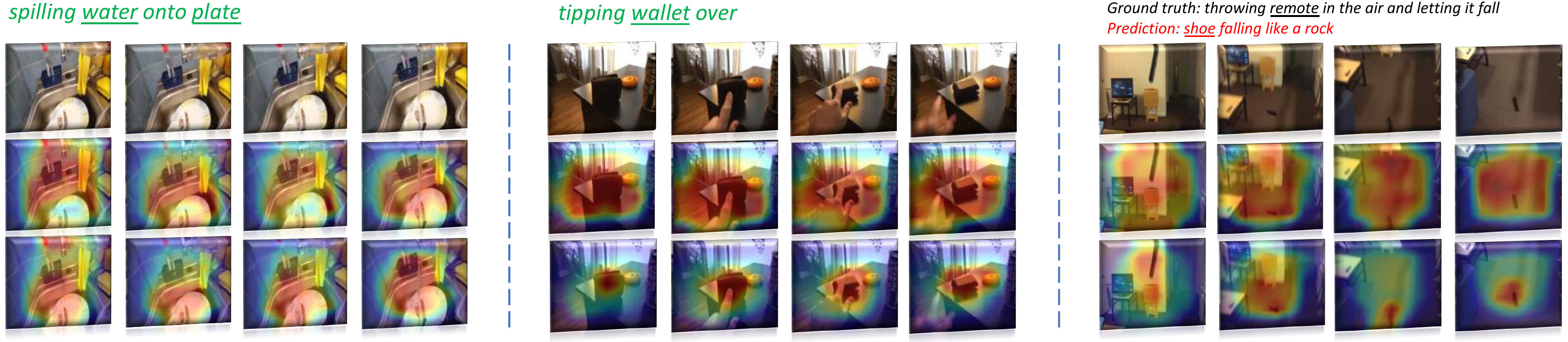}
\caption{\textbf{Examples of attention maps} yielded by the Dual Attention Network with their predicted labels above. For each clip four frames are shown out of eight frames for TRN with a stride of two. The first row is the input frames while the second and third ones are attention maps for recognizing \emph{action} and \emph{object} respectively. The model accurately learns the alignment between actions and objects, even when the background is complicated (\eg, $1^{st}$ clip), or the predicted labels are wrong (\eg, $3^{rd}$ clip). These examples are drawn from validation subsets.}
\label{fig:visualization}
\vspace{-0.1cm}
\end{figure*}

\subsection{Representing videos with neural networks}
\label{sec:extract_video_representation}
There are two \emph{de facto} paradigms to extract video representations: 1) \emph{Image-based} models which use spatial convolutional kernels to process frames independently, and later perform temporal feature aggregation by another model such as a Long Short-Term Memory network (LSTM)~\cite{hochreiter1997long} or a Temporal Relation Network (TRN)~\cite{Zhou_2018_ECCV}; 2) \emph{Video-based} models which apply convolutional kernels across frames to process a video with spatial and temporal dimensions directly.\footnote{We group the models based on the \emph{domain} that they use for extracting features of each frame, \ie, whether cross-time dynamics are exploited, rather than types of convolutional kernels.}

{\bf Image-based models.} 
Given a video $V$ with $T$ frames, CNN features from each frame are extracted \emph{independently}, resulting in a set of $T$ raw features $\left\{ \bm{f}_1, \bm{f}_2, \cdots, \bm{f}_T \right\}$, where $\bm{f}_k \in \mathbb{R}^{d\times{}N}$, $d$ is the feature dimension and $N=HW$ is the vectorized spatial dimension of the feature map. The CNN features are then averaged by global pooling over the spatial dimension, \ie,
\begin{equation} 
    \bm{\bar{f}}_k = \frac{1}{N}\sum_{i=1}^{N}{\bm{f}_k\left[ ., i \right]}
\end{equation}

After that, various modules that process and fuse information across temporal domain can be applied on top of extracted features. For example, all frames can be modeled by an LSTM, resulting in the final representation of a video $\hat{\bm{v}}$ as:
\begin{equation}
    \hat{\bm{v}} = \mathrm{LSTM}\left( \bm{\bar{f}}_1, \bm{\bar{f}}_2, \cdots, \bm{\bar{f}}_T\right)
\end{equation}

Alternatively, TRN~\cite{Zhou_2018_ECCV} is a simple yet effective network module recently proposed to explicitly learn and model temporal dependencies across sparsely sampled frames at different temporal scales. TRN can be applied on top of any 2D CNN architecture. More specifically, an \emph{n}-order relation, for a given number $n$, is modeled as:
\begin{equation} \label{eqn:TRN}
    R_n(V) = h_{\phi}{\left( \sum_{k_1 < k_2 < \cdots < k_n}{g_\theta{( \bm{\bar{f}}_{k_1}, \bm{\bar{f}}_{k_2}, \cdots, \bm{\bar{f}}_{k_n})}} \right)}
\end{equation}

Here $h_\phi$ and $g_\theta$ are both multi-layer perceptrons (MLPs) fusing features of different frames. For the sake of efficiency, rather than summing over all possible choices of $n$ ordered frames, a small number of tuples uniformly sampled are chosen. The model can be extended to capture relations at multiple temporal scales by considering different values of $n$. The final representation of a video is an aggregation of a \emph{2}-order TRN up to an \emph{n}-order TRN:
\begin{equation}
    \hat{\bm{v}} = R_2(V) + R_3(V) + \cdots + R_n(V)
\end{equation}
where $n$ is a hyperparameter of the model.

{\bf Video-based models.} A video-based model operates on multiple frames within a video. As a result, given a video with $T$ frames, features of each frame are not independent anymore, so that cross-time dynamics may be learned in this way. Besides, temporal down-sampling is often adopted to form a sufficiently large receptive field over temporal domain, so that the number of remaining frames $T'$ is less or equal than $T$. Denote a set of $T'$ features as $\left\{ \bm{f}_1, \bm{f}_2, \cdots, \bm{f}_{T'} \right\}$, where $\bm{f}_k$ can be a super frame if $T' < T$, and like in image-based models each frame is then averaged by spatial global pooling. Since dynamics are expected to be learned implicitly within convolutional neural networks, the final representation of a video is usually acquired by averaging across all (super) frames:
\begin{equation} \label{eqn:3dmodel}
    \hat{\bm{v}} = \frac{1}{T'}\sum_{k=1}^{T'}{\bm{\bar{f}}_k}
\end{equation}

\subsection{Object recognition in videos}
\label{sec:object_recognition}
Given a video of an HOI, we want to recognize the action and the object associated with the interaction. For example, for a ``\emph{playing}" action, we want our model to recognize that it is ``\emph{playing a violin}" rather than ``\emph{playing a piano}". For training we assume that labels are provided at the video level without bounding-boxes. A straight-forward method for joint action-object recognition is to add a separate classification head for object recognition alongside the head for action recognition. Note that the task differs from the standard object recognition in static images, because the model should look for the objects being manipulated by the actor instead of those in the background. As a result, the object head should also utilize feature representations containing temporal information, \ie, for image-based models such as TRN, another multi-scale TRN module is used for object recognition, whereas for video-based models we simply use another MLP.

\subsection{Dual attention module}
\label{sec:dual_attention_module}
We propose a dual attention model for action and object recognition, as illustrated in Figure~\ref{fig:att_block}. The model is not dependent on a specific CNN architecture thus it is general and extensible. The dual attention uses action priors to attend image features for objects, and object priors for actions. Suppose that we have the probabilities $\bm{p}^a$ and $\bm{p}^o$ over actions and objects respectively of their likelihood to appear in the video. First, we apply two MLPs to encode these probability vectors into two intermediate feature representations $\bm{h}^a, \bm{h}^o \in \mathbb{R}^d$. The dual attention module takes input of the visual features at each frame and generates the object and action attention distributions over $N$ regions of each frame:
\begin{equation}
    \bm{z}_k^a = \bm{w}_a^T\mathrm{tanh}\left( \bm{W}_a\bm{f}_k + \bm{W}_{ota}\bm{h}^o\mathbbm{1}^T \right)
\end{equation}
\begin{equation}
    \bm{z}_k^o = \bm{w}_o^T\mathrm{tanh}\left( \bm{W}_o\bm{f}_k + \bm{W}_{ato}\bm{h}^a\mathbbm{1}^T \right)
\end{equation}
\begin{equation}
    \bm{\alpha}_k = \mathrm{softmax}(\bm{z}_k^a)
\end{equation}
\begin{equation}
    \bm{\beta}_k = \mathrm{softmax}(\bm{z}_k^o)
\end{equation}
where $\mathbbm{1} \in \mathbb{R}^N$ is a vector whose elements are all equal to 1. $\bm{W}_a, \bm{W}_o \in \mathbb{R}^{N\times{}d}$ and $\bm{w}_a, \bm{w}_o \in \mathbb{R}^{N}$ are the weights to be learned. $\bm{W}_{ota}, \bm{W}_{ato} \in \mathbb{R}^{N\times{}d}$ are parameters for object-to-action attention and action-to-object attention, respectively. $\bm{\alpha}_k, \bm{\beta}_k \in \mathbb{R}^{N}$ are the attention weights over spatial features in $\bm{f}_k$. The representation of each frame is obtained by a weighted-average over its spatial domain:
\begin{equation}
    \tilde{\bm{f}}_k^a = \sum_{i=1}^{N}{\alpha_{k,i} \bm{f}_k\left[ , i \right]}
\end{equation}
\begin{equation}
    \tilde{\bm{f}}_k^o = \sum_{i=1}^{N}{\beta_{k,i} \bm{f}_k\left[ , i \right]}
\end{equation}
Finally, for $x$ in $\{ a, o\}$, we obtain representations of a video for action and object respectively by substituting $\bar{\bm{f}}_k^x$ with $\tilde{\bm{f}}_k^x$ in Equation~\ref{eqn:TRN} or~\ref{eqn:3dmodel}.

\begin{table*}[t]\centering
\subfloat[Joint learning of two tasks. Dual attention is better than multi-task learning at exploiting action and object information. \label{tab:ablation:multi_task}]{
\tablestyle{2pt}{1.05}
\begin{tabular}{l|x{28}x{28}|x{28}}
\multicolumn{1}{c|}{method} & top-1(A) & top-5(A) & top-1(O)\\
\shline
baseline & 44.6 & 73.9 & 58.2 \\
\hline
multi-tasking & 45.7 & 75.0 & 59.9 \\
dual attention & \textbf{46.6} & \textbf{75.6} & \textbf{60.1} \\
 \multicolumn{4}{c}{~}\\
\end{tabular}}\hspace{3mm}
\subfloat[Attention guided by ground-truth labels. The significant improvements indicate that actions and objects are indeed closely intertwined.  \label{tab:ablation:gt_attention}]{
\tablestyle{3pt}{1.05}
\begin{tabular}{l|x{28}x{28}|x{28}}
\multicolumn{1}{c|}{method}  & top-1(A) & top-5(A) & top-1(O)\\
\shline
baseline & 44.6 & 73.9 & - \\
& - & - & 58.2 \\
\hline
GT-object att. & \textbf{50.2} & \textbf{79.7} & -  \\
GT-action att. & - & - & \textbf{67.0} \\
\end{tabular}}\hspace{3mm}
\subfloat[Self attention vs. dual attention. Action and object priors offer a better attention mechanism for recognition.   \label{tab:ablation:self_att}]{
\tablestyle{3pt}{1.05}
\begin{tabular}{l|x{28}x{28}|x{28}}
\multicolumn{1}{c|}{method}  & top-1(A) & top-5(A) & top-1(O)\\
\shline
baseline & 44.6 & 73.9 & 58.2 \\
\hline
self attention & 45.3 & 74.4 & 58.3  \\
dual attention & \textbf{46.6} & \textbf{75.6} & \textbf{60.1} \\
 \multicolumn{4}{c}{~}\\
\end{tabular}}
\vspace{-.5em}
\caption{Ablation study. A: action recognition; O: object recognition. The baseline is a TRN-4 network.}
\label{tab:ablations}
\vspace{-0.2cm}
\end{table*}

\subsection{Full architecture}
\label{sec:full_framework}
The full architecture is illustrated in Figure~\ref{fig:dual_attention_network}. 
Given a video, the network first predicts the plausible action and object labels using two separate heads. The prediction results serve as the priors of actions and objects, which are subsequently used to produce the attention maps for objects and actions, via the dual attention module. A second prediction is performed with the attention-based enhanced features. This two-step scheme expresses the interaction between human and objects. The two prediction modules along with the attention module are integrated into one network for end-to-end learning. Some actions may involve multiple entities, for instance, ``\emph{put something on something}". We therefore use two softmax classifiers to predict the objects. If the order of the objects is exchangeable, \eg, in the category ``\emph{move something and something closer}", 
the classifiers will learn to predict the objects in the order as they appear in the ground truth to avoid ambiguity.
We use a \emph{null} label as a placeholder for those action classes with only a single object. 

\section{Experiments}
We conduct comprehensive experiments below to validate
the efficacy of our proposed Dual Attention Network.


\definecolor{demphcolor}{RGB}{144,144,144}
\newcommand{\demph}[1]{\textcolor{demphcolor}{#1}}
\begin{table*}[!tbp]
\centering
\small
\tablestyle{4pt}{1.05}
\begin{tabular}{l|l|l|l|l|x{30}x{30}|x{30}x{30}}
\multicolumn{1}{c|}{model} & \multicolumn{1}{c|}{backbone} & \multicolumn{1}{c|}{domain} & \multicolumn{1}{c|}{modality} & \multicolumn{1}{c|}{frames}& top-1 val & top-5 val & top-1 test & top-5 test \\
\shline
\multirow{1}{*}{TSN$^\dagger$~\cite{wang2016temporal}} & BN-Inception & 2D & RGB & 8 & 41.1 & 69.3 & - & - \\
\hline
TSN Dual Attention [ours] & BN-Inception & 2D & RGB & 8 & 42.1 & 71.2 & - & - \\
\hline
I3D$^\dagger$~\cite{carreira2017quo} & ResNet-50 & 3D & RGB & 16 & 43.8 & 73.2 & - & - \\
\hline
2D-CNN w/ LSTM~\cite{mahdisoltani2018fine} & VGG-like & 2D & RGB & 48 & 40.2 & - & 38.8 & - \\
3D-CNN w/ LSTM~\cite{mahdisoltani2018fine} & VGG-like & 3D & RGB & 48 & 51.9 & - & 51.1 & - \\
2D-3D-CNN w/ LSTM~\cite{mahdisoltani2018fine} & VGG-like & 2D + 3D & RGB & 48+48 & 51.6 & - & 50.4 & - \\
\hline
TSM$^{\ddagger}$~\cite{lin2018temporal} & ResNet-50 & 3D & RGB & 8 & 56.7 & 83.7 & - & - \\
TSM$^{\dagger}$ & ResNet-50 & 3D & RGB & 8 & 54.0 & 81.3 & - & - \\
\hline
TSM Dual Attention [ours] & ResNet-50 & 3D & RGB & 8 & 55.0 & 82.0 & - & - \\
\hline
\multirow{2}{*}{TRN~\cite{Zhou_2018_ECCV}} & BN-Inception & 2D & RGB & 8 & 48.8 & 77.6 & 50.8 & 79.3  \\
 & BN-Inception & 2D & RGB + Flow & 8+8 & 55.5 & 83.0 & 56.2 & 83.1 \\
\hline
\multirow{2}{*}{TRN Dual Attention [ours]} & BN-Inception & 2D & RGB & 8 &  51.6 & 80.3 & 54.0 & 81.9 \\
 & BN-Inception & 2D & RGB + Flow & 8+8 & 58.4 & 85.2 & 60.1 & 86.1 \\
\end{tabular}
\caption{Comparisons to state-of-the-art methods on Something-V2, with results on both the validation and test subsets. $^\dagger$: Our re-implemented model. $^{\ddagger}$: From original paper, pre-trained on Kinetics~\cite{kay2017kinetics} asides from ImageNet~\cite{deng2009imagenet}.}
\label{tab:sthsth_sota}
\end{table*}

\subsection{Implementation details}
\label{sec:impl_details}
We choose Temporal Segment Networks (TSN)~\cite{wang2016temporal} and TRN~\cite{Zhou_2018_ECCV} as the backbones among image-based models, and Temporal Shift Module (TSM)~\cite{lin2018temporal} among video-based models, given their demonstrated superior performance. We did not choose I3D~\cite{carreira2017quo} because the temporal down-sampling rate is overly large (\eg, 16 frames input and 2 super frames output) so that it is naturally inproper to use spatial attention.

\vspace{-0.3mm}
{\bf Base networks.} Following~\cite{Zhou_2018_ECCV} and~\cite{lin2018temporal}, we adopt Inception with Batch Normalization~\cite{Ioffe2015batchnorm} (BN-Inception) as our base models of TSN and TRN, while ResNet-50~\cite{he2016deep} pre-trained on ImageNet~\cite{deng2009imagenet} as it of TSM for fair comparisons. The input size is set to $224{\times}224$. The spatial size of output features is $7{\times}7$ with $1024$ and $2048$ channels for BN-Inception and ResNet-50, respectively. We append dropout~\cite{srivastava2014dropout} after the extracted features, with a ratio of $0.5$.

\vspace{-0.3mm}
{\bf Dual attention module.} The dual attention module generates distributions over the spatial grids of feature maps for each frame. To embed the probabilities of action or object labels, we use a two-layer MLP with ReLU activations~\cite{nair2010rectified}. Both layers of the MLP have $512$ channels. We project image features into $512$ channels by a single-layer perceptron before feeding them into the attention module.

\vspace{-0.3mm}
{\bf Recognition heads.} We use the same sampling strategy as in~\cite{Zhou_2018_ECCV} for multi-scale TRNs. $g_\phi$ is a two-layer MLP with $256$ units per layer, while $h_\phi$ is a two-layer MLP whose output channels match the number of classes. We do not use dropout within classification heads. For both TSN and TSM the recognition head is a two-layer MLP. The classification heads do not share weights between the first (pre-attention) prediction and the second (post-attention) prediction. This design does not introduce computational overhead as the CNN feature extraction, which dominates the computation, is shared.

\subsection{Setup}
\label{sec:exp_preliminaries}
{\bf Dataset.}
We use Something-something dataset V2~\cite{goyal2017something}, a video action dataset for human-object interactions, with $220,847$ videos from $174$ classes. Those classes are fine-grained so a model needs to distinguish actions such as ``lifting up \emph{one end of} something then letting it drop down'' from actions such as ``lifting something up \emph{completely} then letting it drop down''. This requires the model to look into details of different actions. Note that object labels (\emph{nouns}), provided in V2 by the workers, may result in some inconsistencies: a mobile phone can be depicted as \emph{``phone''}, \emph{``mobile phone''}, \emph{``a phone''}, \emph{``a black phone''} or even \emph{``iPhone''}. We therefore merge nouns describing the same or similar objects, for a total of $307$ object clusters (see supplementary material for details). We conduct our study on this dataset because it is among the very few ones containing videos of diverse human-object interactions instead of a few pre-defined relations between actions and objects. Also it is one of a few large-scale video dataset which provides object labels.

{\bf Training details.} We use a multi-scale 4-frame TRN (TRN-4) in all our ablation study for efficiency. Results of an 8-frame TSN (TSN-8), an 8-frame TRN (TRN-8) and an 8-frame TSM (TSM-8) are included in the final experiment. The networks are trained end-to-end. We augment the data during training by scale and aspect-ratio jittering. The batch size is set to 32 for TRN-4, 16 for TSN-8 and TRN-8, and 8 for TSM-8 due to GPU memory limitation. We train all models on a server with 8 GPUs for 70 epochs. It starts with a learning rate of $0.01$, and is reduced by a factor of $10$ at epoch 50 and 65. We use a momentum of $0.9$. The weight decay of models with BN-Inception is set as $0.0001$, whereas $0.0005$ for models with ResNet. We train our models with unfrozen Batch Normalization, which effectively stabilizes the training procedure.

\subsection{Main results}
\label{sec:main_results}

Results on the validation subset are in Table~\ref{tab:ablation:multi_task}. Our dual attention model attached on a TRN-4 network yields accuracies of $46.6/75.6$ (top1/top-5) on action recognition and of $60.1$ (top-1) on object recognition, a $2.0/1.7$ and $1.9$-point boost over the baseline. Compared to a separately trained model of joint learning of actions and objects (multi-tasking), our approach achieves superior performance, indicating that dual attention is a better approach to utilize the action and object information interchangeably.

Figure~\ref{fig:visualization} visualizes attention maps learned by our model. For each clip, the first row contains four out of the eight frames chosen by the TRN module. The second and third rows are attention maps for \emph{actions} and \emph{objects}. We see that our model learns meaningful alignment between actions and objects. For action recognition, the attention map generally covers a larger space capturing the global information of the entire action series; for object recognition, the attention map is sharp and neat, mostly on the object being manipulated by the actor. Surprisingly, the model can attend to the relevant region and predict correct classes even when the background is complex, \eg, the first example, in which the model finds the dishes in the sink as well as the water being spilled but ignores the background. In cases where the model produces inaccurate predictions, \eg, the third example, our model still looks at reasonable regions across frames, although it seems unable to recognize the fine-grained categories.

{\bf Cohesion of actions and objects.}  
In order to understand the potential maximum performance gain that we can expect from our approach, we experiment with using ground-truth annotations as action and object priors instead of predicting them. As shown in Table~\ref{tab:ablation:gt_attention}, ground-truth guided attentions show a  remarkable improvement for both action recognition ($5.6\%/5.8\%$) and object recognition (8.8\%). This demonstrates that action and object recognition are closely intertwined, and that improving the first prediction in our approach can lead to an even bigger boost to performance. 

\begin{figure}[!tbp]
\centering
\includegraphics[width=0.4\textwidth]{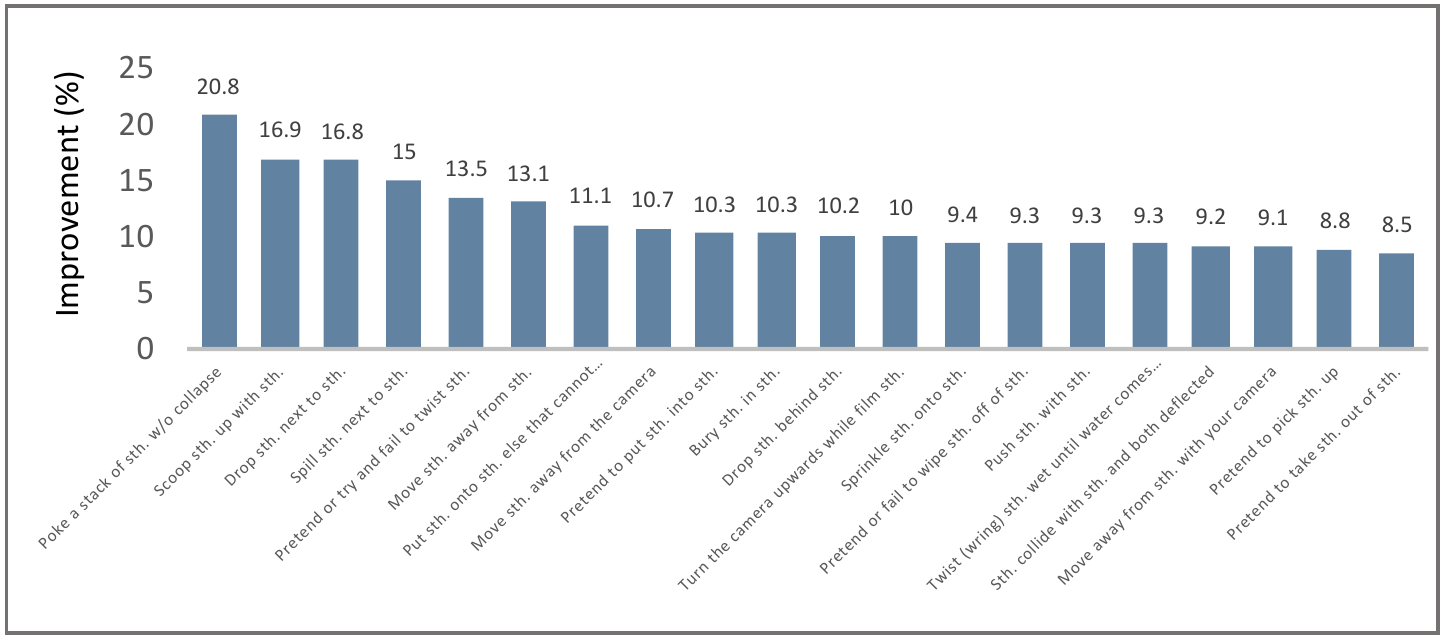}
\caption{\textbf{Class-wise improvements} of the Dual Attention Network with respect to the baseline model. Action classes closely associated with certain objects are improved.}
\label{fig:class_wise_cmp}
\end{figure}

Figure~\ref{fig:class_wise_cmp} shows the class-wise improvements over baseline with dual attention. We can see that the performance of action categories which are closely associated with certain types of objects is boosted. For example, a \emph{liquid} prior is helpful for recognizing \emph{``spill something''}. Similarly, \emph{something untwistable} is helpful for predicting \emph{``pretend or try and fail to twist something''}. Meanwhile, the performance of actions related to the physical localization of objects, such as \emph{``turn the camera upwards while film something''} and \emph{``move something away''} also gets better. The improvement on these action categories strongly indicates that our dual attention mechanism facilitate the model to trace the object manipulated by the actor.

{\bf Self attention vs. Dual attention.}
We train a model by generating attention maps from image features only w/o the guidance of priors, termed self-attention model.
Table~\ref{tab:ablation:self_att} compares our approach with the self-attention one. The inferior performance of self attention suggests that actions and objects priors indeed provide useful information for objects and actions recognition respectively.

\begin{figure*}[!th]
\centering
\includegraphics[width=0.9\textwidth]{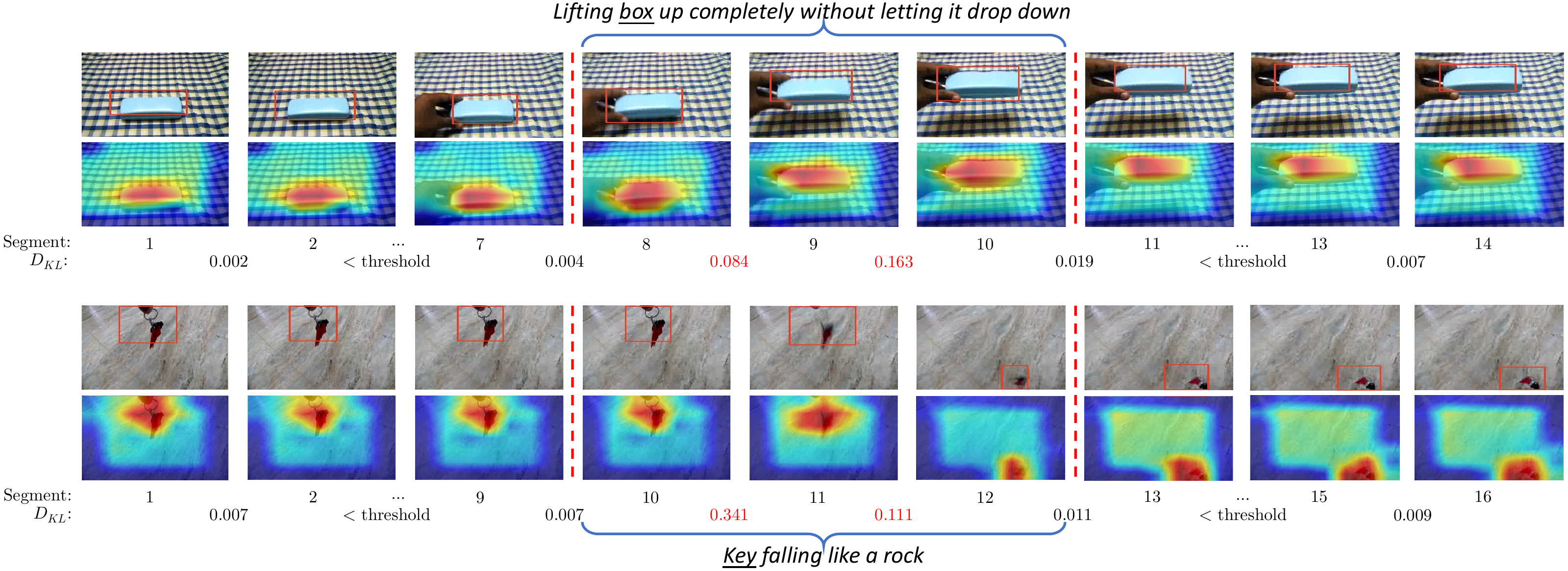}
\caption{\textbf{Visualization of spatial and temporal localization}. We visualize one frame out of each segment (four frames). Our method find the object being manipulated, as well as the segments in which the action actually happens.}
\label{fig:vis_localization}
\vspace{-0.2cm}
\end{figure*}

\subsection{Comparisons with state-of-the-art methods} 
Table~\ref{tab:sthsth_sota} summarizes results on Something-Something V2. We do not use the earlier version as object annotations are missing in V1 and label noise is greatly reduced in the latest release. We compare our approach with TSN~\cite{wang2016temporal}, I3D~\cite{carreira2017quo}, 2D and 3D CNNs with LSTM from Something-Something~\cite{mahdisoltani2018fine}, and previous state-of-the-art TRN and TSM. These approaches differ from each other in many aspects such as backbones, temporal feature fusion techniques, training schemes, number of input frames, model domains and modalities. Still, models with the dual attention module surpass all their counterparts. Specifically, TSN dual attention is better than original TSN by $1.0/1.9\%$ points; TSM dual attention is better than original TSM by $1.0/0.7\%$ points; TRN dual attention achieves top-1 accuracy of $51.6\%$ and top-5 accuracy of $80.3\%$ on the validation subset, which is better than any previous 2D model. We conjecture that performance boost of TRN being larger than it of TSN and TSM is because TRN has more complex recognition heads so that it might be able to better exploit attended features. When TRN dual attention is turned into a two-stream models by adding an optical flow branch in TRN, our approach further boosts the performance to $58.4/85.2\%$. 

\section{Weakly-supervised Localization}
\label{sec:localization}
We can reason about human-object interactions by inspecting attention maps yielded by the model. Here, we apply the model to two weakly-supervised localization tasks: \emph{spatiotemporal localization} and \emph{object affordance segmentation}.

\subsection{Spatiotemporal localization}

The attention maps, learned from video-level action and object labels only, can accurately localize objects in the spatial domain, and actions in the temporal domain. The localization task requires the object attention map of \emph{all} frames available. This is achieved by running the dual attention network alone with predictions of the object and action from our model, and the CNN features extracted from every frame, as input. A single forward pass with a batch of $n$ frames can generate $n$ attention maps.

\begin{table}[!t]
\begin{center}
\begin{tabular}{l|r}
\textbf{Action Category} & \textbf{A.D.} \\
\shline
Piling something up & 49.1 \\
Stacking number of something & 48.4 \\
Pouring sth. into sth. until it overflows & 47.9 \\
Pouring something into something & 43.7 \\
Digging something out of something & 43.5 \\
\hline 
Pretending to put something on a surface & 35.4 \\
Spinning something so it continues spinning & 34.8 \\
\hline
Putting something on a surface & 29.8 \\
Spinning something that quickly stops spinning & 28.5 \\
Tipping something over & 27.7 \\
Uncovering something & 26.8 \\
Something falling like a rock & 25.7 \\
Throwing something onto a surface & 22.7 \\
\end{tabular}
\end{center}
\vspace{-0.2cm}
\caption{The average duration (A.D.) of \emph{trimmed} videos in each action category. The A.D. is measured by frames and the fps rate is 12. The results are in accordance with our human knowledge.}
\label{tab:action_length}
\vspace{-0.1cm}
\end{table}

\begin{table}[!t]
\begin{center}
\begin{tabular}{l|r|r|r}
\textbf{Model} & IoU=0.3 & IoU=0.4 & IoU=0.5 \\
\shline
Dual Attention & 72.5 & 56.0 & 33.7 \\
\hline
Self Attention & 62.2 & 40.4 & 26.4 \\
\end{tabular}
\end{center}
\vspace{-0.2cm}
\caption{Object localization results (in Average Precision) of dual attention and self attention on the \emph{validation} subset.}
\label{tab:spatial_loc_eval}
\vspace{-0.2cm}
\end{table}

{\bf The ``where'': spatial object localization.} 
We generate object bounding boxes by thresholding the object attention map. We set the threshold to $60\%$ of the maximum weight in the map. We then apply the flood-fill algorithm to find the connected regions. Bounding-boxes are generated by calculating the minimum and maximum coordinates of each region. We always take the largest bounding-box as a prediction, while the second largest one (if available) is optionally taken based on its size and the number of predicted objects. 

{\bf The ``when'': temporal action localization.} 
We observe that a large amount of human-object interactions take place once the object starts to move. Thus, we can associate the start and end of an action via the alteration of the attention maps. We divide a video into segments covering $1/3$ second each, \ie, four frames in one segment for videos from the Something-Something dataset. We average the attention maps within each segment to reduce the margin of error. We measure the difference between two object attention maps $P$ and $Q$ (as they are two discrete probability distributions) via the Kullback–Leibler divergence, \ie:
\begin{equation}
    D_{\mathrm{KL}} (P || Q) = -\sum_i{P(i)\log{\frac{Q(i)}{P(i)}}}
\end{equation}
where the sum is over the discrete points in the domain of the distribution. We filter out the leading (trailing) segments if the difference to its preceding (succeeding) segment is below a threshold. The remaining segments are considered as the interval in which an action happens. We set the threshold to $0.06$. We consider an action spanning over the entire video if the filtered video is shorter than one second to avoid actions such as \emph{``holding something''} or \emph{``showing something''}.

{\bf Results.}
We perform temporal and spatial localization on videos from the validation subset (see Figure~\ref{fig:vis_localization}). We can see that the object being interacted is highlighted and a reasonable bounding-box associated is generated accordingly. Due to the stable and accurate attention map, we are able to eliminate leading and trailing frames irrelevant to the action and find the segments wherein the box is being lifted and the key is falling. We note that compared to using optical flow our approach has more advantages that it can be performed with sparsely sampled segments if the video is very long, and is more robust to camera shake.

We conduct quantitative evaluation of weakly-supervised spatial localization by dual attention model and self attention model on the validation subset, as shown in Table~\ref{tab:spatial_loc_eval}. We randomly sample 100 videos from validation subset and annotate 2 random frames in each video. We report the average precision (AP) under various intersection-of-union (IoU) criteria. As can be seen from Table~\ref{tab:spatial_loc_eval}, our dual attention model yields much better localization accuracy than the self-attention model, indicating that action priors help the model better localize the object being manipulated. We further analyze the statistics of the trimmed videos. Out of total 24,777 videos, 16,592 ($\scriptsize{\sim}67\%$) are trimmed by our temporal localization technique. The average trimmed length is 13 frames ($\scriptsize{\sim}1$ second), which, compared to the average length of 3.1 seconds, accounts for $1/3$ of overall frames. After performing temporal localization, we also analyze the average length of videos in each action category and summarize the results in Table~\ref{tab:action_length}. 
The longest actions involve \emph{``piling something up''} and \emph{``stacking number of something''} whereas the shortest ones involve \emph{``throwing something onto a surface''} and \emph{``something falling like a rock''}. It is also interesting to see that \emph{pretending} to do something is longer than actually doing it, something \emph{continues} spinning is longer than it quickly stops spinning, and pouring something until \emph{overflowing} is longer than pouring something. 

\begin{figure}[!t]
\centering
\includegraphics[width=0.84\linewidth]{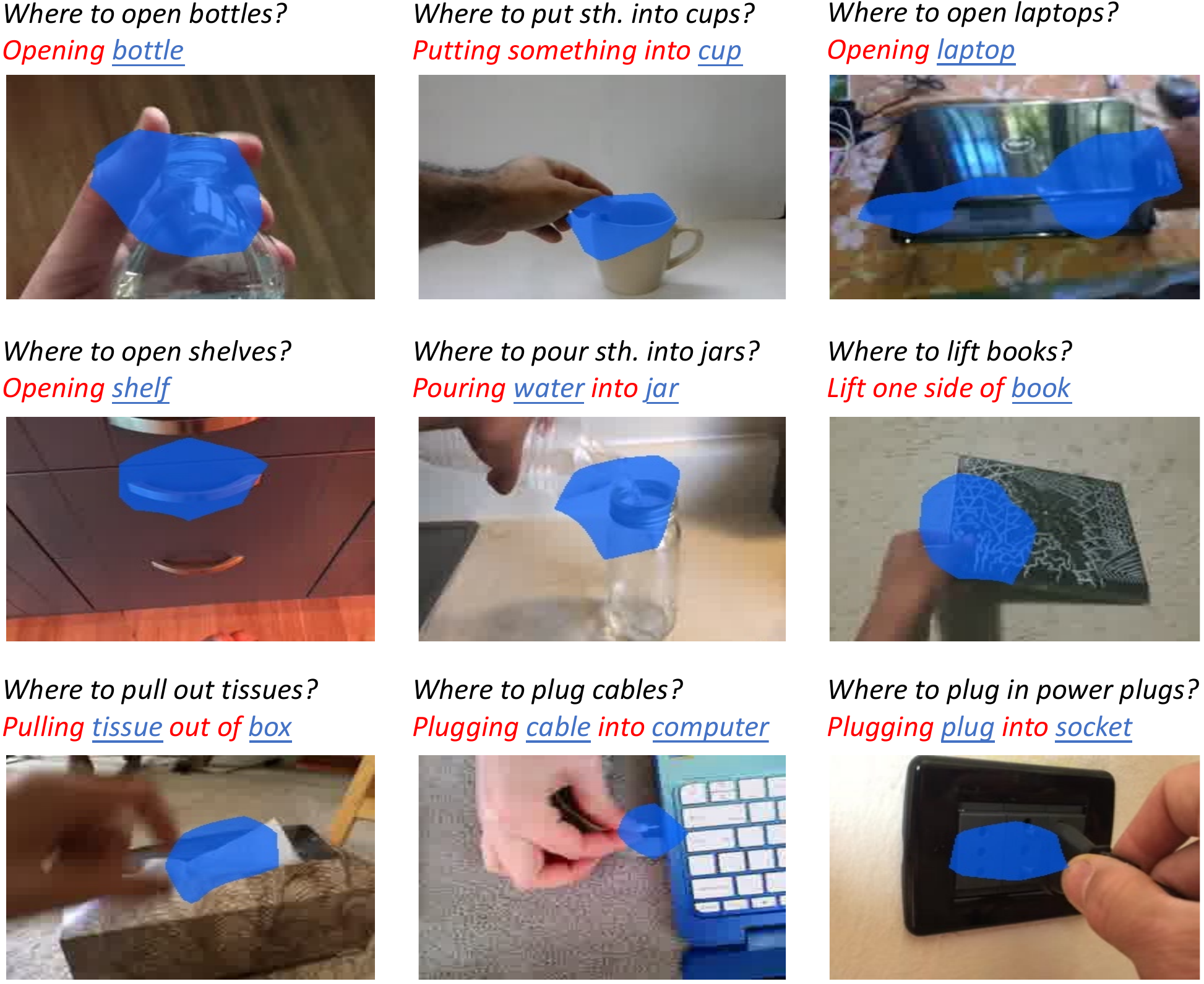}
\caption{\textbf{Examples of object-affordance segmentation}. The model trained with video-level annotations can find object parts associated with possible ongoing actions.}
\label{fig:affordance}
\vspace{-0.2cm}
\end{figure}

\subsection{Object affordance segmentation}
Humans can learn the roles of different parts of an object by observing how the object is being used, \ie, by watching examples of ``pouring water into a bottle'' we can infer not only that water can be poured, but also through which part of the bottle the water can be poured. Our model can learn that detailed information. Given a question such as \emph{``Where to plug cables?''}, we find the videos with related labels, \eg, videos labeled with \emph{``plugging a cable into a computer''}. Note that since we would have acquired the ground-truth object label when retrieving videos, the ground-truth object-guided attention is used to create attention maps; meanwhile, this model is trained with ground-truth objects to perform action recognition thus it mixes actions and objects for affordance discovery. After acquiring the attention maps, we then segment by a threshold of $60\%$ of the maximum attention weight. The results are in Figure~\ref{fig:affordance}: the model focuses on the object parts associated with the action instead of the whole object. For example, the model focuses on the brim of a cup for videos involving \emph{pouring something into a cup} or \emph{putting something into a cup}. Importantly, the model knows to focus on the handle of a shelf even in a still image. This enables us to effectively parse object parts and infer their affordance even when the labels used for training are at the video-level.
\section{Conclusion}
Dual Attention Networks is proposed to recognize human-object interactions. It achieves very competitive performance on Something-Something V2 dataset. Based on actions/objects priors, The model is able to produce intuitive and interpretable attention maps which can enhance video feature representations for improving the recognition of both objects and actions and enable better video understanding.

\vspace{+0.25cm}
\textbf{Acknowledgement:} This work was supported by the MIT-IBM Watson AI Lab, as well as the Intelligence Advanced Research Projects Activity (IARPA) via Department of Interior/ Interior Business Center (DOI/IBC) contract number D17PC00341. The U.S. Government is authorized to reproduce and distribute reprints for Governmental purposes notwithstanding any copyright annotation thereon. Disclaimer: The views and conclusions contained herein are those of the authors and should not be interpreted as necessarily representing the official policies or endorsements, either expressed or implied, of IARPA, DOI/IBC, or the U.S. Government.

{\small
\bibliographystyle{ieee_fullname}
\bibliography{main}
}

\end{document}